\documentclass[authoryear, 5p, twocolumn, 11pt]{elsarticle}
\usepackage{amsfonts, amsmath, bm, amsthm, amssymb}
\usepackage{graphicx}
\usepackage{xcolor}
\usepackage{subcaption}
\usepackage{soul} 

\usepackage{parskip} 
\usepackage{setspace}
\usepackage{booktabs, multirow, makecell}
\usepackage{comment}
\usepackage[font=itshape]{quoting}

\usepackage[font=scriptsize]{caption}
\usepackage[hyphens]{xurl} 
\usepackage{breakurl} 
\usepackage{hyperref}
\hypersetup{
    breaklinks = true,
    urlcolor = black,
	colorlinks = true,
	citecolor = blue, 
	linkcolor = blue 
}

\newcommand{\ra}[1]{\renewcommand{\arraystretch}{#1}} 

\newcommand{\indep}{\perp \!\!\! \perp}

\theoremstyle{highlight}
\newtheorem{openpoint}{Open Point}

\usepackage{natbib}
\defcitealias{Directive200848}{Consumer~Credit~Directive}
\defcitealias{Directive200043}{Directive~2000/43/EC} 
\defcitealias{Directive200078}{Directive~2000/78/EC} 
\defcitealias{Directive2004113}{Directive~2004/113/EC} 
\defcitealias{Directive200654}{Directive~2006/54/EC} 
\defcitealias{EUCharter}{EU Charter of Fundamental Rights}
\defcitealias{EUnondisc}{EU non-discrimination website}
\defcitealias{USCivilRights}{US Civil Rights Act}
\defcitealias{ECOA}{Equal Credit Opportunity Act}
\defcitealias{carson1996}{Carson v. Bethlehem Steel Corp.}
\defcitealias{texdept2015}{Texas Dept. of Housing \& Community Affairs v. Inclusive Communities Project, Inc.}
\defcitealias{castaneda1977}{Castaneda v. Partida}
\defcitealias{ricci2009}{Ricci v. DeStefano}
\defcitealias{USgov}{Blueprint for AI}
\defcitealias{UKgov}{pro-innovation approach}
\defcitealias{SPAINngov}{Comprehensive Law for Equal Treatment and Non-Discrimination}
\defcitealias{ITConst}{Italian Constitution}
\defcitealias{AIAct}{AI Act}
\defcitealias{NationalAICommission}{National AI Commission Act}
\defcitealias{AISafetyInstitute}{AI Safety Institute}
\defcitealias{EqualityAct}{UK Equality Act 2010}

\makeatletter
\def\ps@pprintTitle{%
 \let\@oddhead\@empty
 \let\@evenhead\@empty
 \def\@oddfoot{}%
 \let\@evenfoot\@oddfoot}
\makeatother

\newif\ifrev
 \revtrue
\ifrev

\else

\fi

\usepackage{tikz}
\usetikzlibrary{shapes, decorations, arrows, calc, arrows.meta, fit, positioning}
\tikzset{
    -Latex,auto,node distance =1 cm and 1 cm,semithick,
    state/.style ={ellipse, draw, minimum width = 0.7 cm},
    point/.style = {circle, draw, inner sep=0.04cm, fill},
    bidirected/.style={Latex-Latex,dashed},
    el/.style = {inner sep=2pt, align=left, sloped}
}

\begin{document}

\title{Fair Enough?\\ A map of the current limitations of the requirements to have fair algorithms.}
\author[1]{Daniele~Regoli\fnref{dr}}
\ead{daniele.regoli@intesasanpaolo.com}

\author[1,2]{Alessandro~Castelnovo\fnref{dr}}
\ead{alessandro.castelnovo@intesasanpaolo.com}

\author[1]{Nicole~Inverardi\fnref{dr}}
\ead{nicole.inverardi@intesasanpaolo.com}

\author{Gabriele~Nanino}
\ead{naninogabriele@gmail.com}

\author[1]{Ilaria~Giuseppina~Penco\fnref{dr}}
\ead{ilaria.penco@intesasanpaolo.com}

\address[1]{Data Science \& Artificial Intelligence, Intesa Sanpaolo S.p.A., Italy}

\address[2]{Dept. of Informatics, Systems and  Communication, Univ. of Milano Bicocca, Italy}


\fntext[dr]{The views and opinions expressed are those of the authors and do not necessarily reflect the views of Intesa Sanpaolo, its affiliates or its employees.}

\begin{abstract}
    In recent years, the increase in the usage and efficiency of Artificial Intelligence and, more in general, of Automated Decision-Making systems has brought with it an increasing and welcome awareness of the risks associated with such systems. One of such risks is that of perpetuating or even amplifying bias and unjust disparities present in the data from which many of these systems learn to adjust and optimise their decisions. This awareness has on the one hand encouraged several scientific communities to come up with more and more appropriate ways and methods to assess, quantify, and possibly mitigate such biases and disparities. On the other hand, it has prompted more and more layers of society, including policy makers, to call for fair algorithms. We believe that while many excellent and multidisciplinary research is currently being conducted, what is still fundamentally missing is the awareness that having fair algorithms is \emph{per se} a nearly meaningless requirement that needs to be complemented with many additional social choices to become actionable. Namely, there is a hiatus between what the society is demanding from Automated Decision-Making systems, and what this demand actually means in real-world scenarios. In this work, we outline the key features of such a hiatus and pinpoint a set of crucial open points that we as a society must address in order to give a concrete meaning to the increasing demand of fairness in Automated Decision-Making systems.
\end{abstract}

    
\maketitle


\section*{Introduction}

Artificial Intelligence~(AI) is experiencing an epoch of unparalleled accomplishments. In particular, Machine Learning~(ML) approaches to AI ---i.e., methods based on mathematical and statistical tools that enable automated systems to learn patterns from data--- are fostering the development of numerous fruitful applications in an incredibly broad range of fields, including computer vision, predictive analytics, and natural language processing.

In the recent years, increasing attention is being devoted to the risks that AI and, more broadly, Automated Decision-Making~(ADM), inevitably brings with it. One of such risks is that ADM systems, and ML approaches in particular, are prone to learning from data a wide range of historical and social biases, and to repeating and amplifying them at a scale that is challenging to monitor and regulate.
With the term fair-AI, or fair-ML, the literature usually denotes the relatively recent area of research committed to study how to assess and correct biases involved in algorithmic decision-making processes. The 2023 Stanford AI Index report~\citep{HAIindex2023}, on the basis of data about paper submissions to top-level conferences, says ``Interest in AI ethics continues to skyrocket''.

Along with this raising attention from the academic world, concerns for the increasing use and diffusion of AI and ADM systems in many industrial sectors emerged from ever broader layers of society. The issue of ethics in AI started gaining popularity, usually under umbrella terms such as ``Trustworthy AI'' or ``Responsible AI''. In AI-related contexts, several \emph{suitcase words} began to be widely used, such as ``bias'', ``fairness'', ``interpretability'', etc.~\citep{lipton2018troubling}. 

More concretely, the 2024 Stanford AI Index report~\citep{HAIindex2024}, based on AI Incident Datbase,\footnote{\href{https://incidentdatabase.ai/}{incidentdatabase.ai}.} highlights that the number of reported incidents related to the ethical misuse of AI is rapidly rising.

Against this background, in the spring of 2021 the European Commission published the ``Proposal for a Regulation of the European Parliament and of the Council laying down harmonised rules on Artificial Intelligence''~\citep{EUproposal}, the very first legal framework for AI. The legislative process has recently concluded with the publication of the Artificial Intelligence Act in the Official Journal of the European Union~\citep{AIAct}.  Since the European proposal, several other countries started discussing and working on regulatory frameworks for AI systems: the \citetalias{USgov} and the \citetalias{NationalAICommission} in the U.S., the \citetalias{UKgov} and the \citetalias{AISafetyInstitute} in the U.K., to name a few. The 2024 Stanford AI Index report~\citep{HAIindex2024} records 148 bills related to AI passed into law throughout the world in the period 2016-2023.

As a result, public and private actors that employ AI and ADM in their processes are driven to attempt to satisfy this requirement for ethics and justice.   
Practitioners and researches have access to several ``fairness toolkits''~\citep[][etc.]{aif360-oct-2018, weerts2023fairlearn, wexler2020whatif}, with the aim of assessing and mitigating unjust disparities in algorithmic outcomes. Unfortunately, the level of maturity and consistency of such toolkits appears to be still very low, and their effectiveness in addressing the overall challenge of ``fairness'' is debated~\citep{lee2021thelandscape}.

Recall that the rising attention and the lively debate that we are today experiencing around the topic of fairness in ADM bear interesting similarities with the ``fairness movement'' in academic research that followed the approval of the \citetalias{USCivilRights} in 1964, as beautifully discussed by \citet{hutchinson201950years}. In particular, the decade after the approval of the Act saw a boom of research into how to mathematically measure unfair discrimination in standardised tests within the educational and employment communities, often with a focus on race. By the mid 1970s, researchers came up with multiple statistical measures of fairness and their mutual incompatibility, in striking resemblance to the current quantitative definitions and the related debate. Unfortunately, by the late 1970s, that movement had mostly disappeared, probably due in part to a general lack of consensus and clarity over standards to determine what constitutes unfair discrimination and, more importantly, over the ethical and legal principles that underpin fairness~\citep{cole2001new}. An issue that even the current ``reboot'' of quantitative research on fairness has not adequately addressed, yet.

\subsection*{Contribution}

This work aligns with a relatively recent stream of research that is focused critically on the general topic of fairness in ADM~\citep{Ruggieri_Alvarez_Pugnana_State_Turini_2023, saxena2023missed, buyl2022inherent, dolata2022asociotechicalview, selbst2019, cooper2021, hoffmann2019where}. The critic is not on the topic \emph{per se} ---whose importance is not disputed--- but rather on some usually overlooked subtleties and assumptions that often lead to overreliance and misplaced trust, which can in turn effectively lead to a deterioration of trust in ADM in the long term. 

Among the generic risks of ``blindly'' embracing simplified recipes, we can cite the so-called \emph{Automation Bias}, namely the propensity to place unmotivated trust on automated decisions, or ---worse--- the possibility of cherrypicking certain simple approaches promoting the false perception that an ADM system respects ethical values. \citet{aivodji2019fairwashing} has evocatively named the latter \emph{Fairwashing}.\footnote{More precisely, \citet{aivodji2019fairwashing} coined the term `fairwashing' with respect to black-box interpretability techniques used to promote false perception of compliance with ethical values, but the concept has a straightforward extension to fair-AI techniques.}

Even if most of the ambiguities and attention points that we here discuss have already been introduced separately by other authors, on the one hand, we try to give an overall perspective, grounding such ambiguities on few foundational intersections of the legal, ethical, and algorithmic perspectives; on the other hand, we approach the topic as a call for action, placing the focus on the fact that most of the ambiguities are a matter of decisions that are not technical in nature, but rather societal, and lie at the intersection of very diverse disciplines. 
In fact, the main goal of this work is to pinpoint a set of open points that constitute obstacles both for researchers in the field of AI and for practitioners and developers of ADM systems to meet the societal requirement of having fair algorithms.

We primarily adopt the viewpoint of practitioners and developers of ADM systems, who must deal with the practical difficulties of identifying potential instances of unfair discrimination in real-world situations and possibly mitigating them when appropriate.

\subsection*{Paper outline}

Even if it is to some extent debatable, we can reasonably agree that the overall perimeter of \emph{unjust} discrimination is group specific. By this, we mean that we regard discrimination as unjust with respect to some specific dimensions, or attributes, that we as a society consider to be somehow protected. Therefore, making decisions involving those attributes is considered morally not justifiable. Typical examples of such attributes are gender, ethnicity, religion, etc.

Under this assumption, we build our critical analysis of fair-AI by distinguishing two broad aspects: 

\begin{enumerate}
    \item \textbf{Sensitive attributes choice}: It is not that discrimination \emph{per se} is unjust, only discrimination with respect to some attributes that we have either to list and agree upon or to define with some reasonable criterion. (see \autoref{sec:P1})

    \item \textbf{What do we mean by unfair discrimination?}: We need to clarify what we mean by making decisions involving such attributes, and in what cases such decision-making represent an unjust discrimination. (see \autoref{sec:P2})
\end{enumerate}
Point 1 underlies several difficulties about what attributes we should monitor and what we should do with subgroups and intersections with respect to those attributes. We devote \autoref{sec:P1} to a discussion of such difficulties. The second point underlies ambiguities that arise when examining the mechanisms the we should consider to be ethically unacceptable by which a sensitive attribute can influence a decision. Section~\ref{sec:P2} is focused on discussing such aspects.

In order to underline the cornerstones of our analysis, we pinpoint a number of \emph{Open Points}: open critical issues that must be addressed and resolved in order to concretely satisfy the need of fair ADM systems.


Finally, we devote \autoref{sec:causality} to analyse the importance of causality when dealing with discrimination, and dicuss about its limitations.

In Table~\ref{tab:summary}, we summarise the key points raised throughout the paper.  

\begin{table*}
\caption{\textbf{Schematic summary} of current open points of fairness requirements for ADM.}
\label{tab:summary}
\centering
\ra{1.2}
\begin{tabular}{@{}lcc@{}}
  \toprule
  
  & \phantom{abc} & \textbf{Open Point} \\
  \midrule

    \multirow{4}{*}{Sensitive attributes choice} && \ref{a:protected_groups}: Identification of protected groups\\
    && \ref{a:sensitive_data_collection}: Sensitive data collection\\
    && \ref{a:group_aggregation}: Group aggregation\\
    && \ref{a:intersectional_bias}: Intersectional bias\\
    \cmidrule{2-3}

    \multirow{6}{*}{What do we mean by unfair discrimination?} && \ref{a:direct_and_indirect_discrimination}: Direct and indirect discrimination\\
    && \ref{a:business-needs}: Objective justification\\
    && \ref{ap:overreliance_on_observational_metrics} Over-reliance on observational metrics\\
    && \ref{a:threshold}: The threshold problem\\
    && \ref{ap:assumptions_of_causal_structure} Assumptions of causal structure\\
    && \ref{ap:use_of_human_attributes_as_causes}: Use of human attributes as causes\\
  \bottomrule

\end{tabular}
\end{table*}

\section{Sensitive attributes choice}
\label{sec:P1}

\subsection{When a given attribute should be \emph{protected}?}

High level non-discrimination principles can be found in several legislative frameworks.  
At the European level, the starting point of analysis on this topic is the \citetalias{EUCharter}. In particular, art~21,1~reads: 
\begin{quoting}
    Any discrimination based on any ground such as sex, race, colour, ethnic or social origin, genetic features, language, religion or belief, political or any other opinion, membership of a national minority, property, birth, disability, age or sexual orientation shall be prohibited.
\end{quoting}
It is worth noting that the EU Charter is addressed to institution and bodies of the European Union and to the Member States, and not to the private sector, at least not directly.
To give concrete application to these principles, various directives have been adopted over time.
They detail provisions for specific protected groups and/or specific domains, such as work, environment, or access to goods and services. See the \citetalias{EUnondisc} and \cite{WACHTER2021why} for more details, and Table~\ref{tab:directives} for a schematic overview. 

Moreover, there are domains where the EU Charter is explicitly referred to as a source of high level principles to be observed, but without any concrete details about their implementation. An example is the \citetalias{Directive200848}, which claims:
\begin{quoting}
    This Directive respects fundamental rights and observes the principles recognised in particular by the Charter of Fundamental Rights of the European Union. In particular, this Directive seeks to ensure full respect for the rules on protection of personal data, the right to property, non-discrimination, protection of family and professional life, and consumer protection pursuant to the Charter of Fundamental Rights of the European Union.
\end{quoting}
The European Commission has in place a proposal of revision of the Directive on consumer credits~\citep{EUproposalCredit} that contains the following Article~6, explicitly on non-discrimination:
\begin{quoting}
    Member States shall ensure that the conditions to be fulfilled for being granted a credit do not discriminate against consumers legally resident in the Union on ground of their nationality or place of residence or on any ground as referred to in Article 21 of the Charter of Fundamental Rights of the European Union, when those consumers request, conclude or hold a credit agreement or crowdfunding credit services within the Union.
\end{quoting}
which seems to suggest that all attributes referred to in the \citetalias{EUCharter} are to be considered sensitive for credit access purposes.

Something not dissimilar can be found in the US legislation. We refer to \citet[chap~6]{barocas-hardt-narayanan} and \citet{barocas2016big} for more details.

\begin{table*}
\caption{\textbf{Schematic summary} of protected categories explicitly covered by EU Directives on non-discrimination. See also the \citetalias{EUnondisc}.}
\label{tab:directives}
\centering
\ra{1.8}
\resizebox{\textwidth}{!}{
\begin{tabular}{@{}lcccccc@{}}
  \toprule
  
  \textbf{Directive} & \phantom{abc} & \textbf{year} & \phantom{abc} & \textbf{Domain of application} & \phantom{abc} & \textbf{Protected categories}\\
  \midrule

    \makecell[l]{Race Equality Directive\\\citepalias{Directive200043}} && 2000 && \makecell[c]{employment, social protection,\\ healthcare, education,\\ access to and supply of goods and services\\ which are available to the public} && race and ethnic origin\\[1ex]
    \makecell[l]{Employment Directive\\\citepalias{Directive200078}} && 2000 && working environment && \makecell[c]{religion or belief,\\ disability, age, sexual orientation}\\[1ex]
    \makecell[l]{Gender Access Directive\\\citepalias{Directive2004113}} && 2004 && \makecell[c]{access to and supply of\\ goods and services} && gender\\[1ex]
    \makecell[l]{Gender Equality Directive\\\citepalias{Directive200654}} && 2006 && employment && gender\\    
  
  \bottomrule

\end{tabular}
}
\end{table*}

In general, we can say that there are high level principles requiring non-discrimination with respect to several dimensions and some more specific legislation regulating single domains and/or single sensitive attributes (such as gender or ``race''), but besides that, there is ambiguity regarding what individual characteristics should be considered as protected or sensitive.

We can summarise the above with the following:
\begin{openpoint}[Identification of protected groups]
\label{a:protected_groups}
    Given a specific phenomenon, what are the groups of people that we should consider as \textbf{protected}, and with respect to which we therefore have to take care of assessing and avoiding any unjust discrimination? 
\end{openpoint}

It is important to point to the \citetalias{EqualityAct} as a noteworthy exception, that explicitly list a set of protected characteristics (namely: age, disability, gender reassignment, marriage and civil partnership, pregnancy and maternity, race, religion or belief, sex, sexual orientation) and addresses them within an organic framework of anti-discrimination legislation.

\subsection{Sensitive data collection}

Suppose that your team is in charge of developing a Decision System to rank applicants to a job posting in order to prioritise interviews. The data you collect come entirely from submitted CVs and the online application form, where applicants are asked questions on standard personal data (name, gender, address, date and place of birth), previous work experience, education, and skills.
While developing the system, you take particular care of avoiding unjustified dependencies of the outputs on information such as gender and nationality of the applicants.
You think you have done all that was possible to prevent any form of unjust discrimination, but when you present your work to your boss, she points out that you actually have no control whatsoever about discrimination with respect to a lot of other sensitive information ---such as political and religious opinion--- for the very trivial fact that you don't have such information to begin with. 
It is well-known that Fairness Through Unawareness, i.e. being blind to sensitive information, is in general not enough to prevent unjust discrimination, or at least some forms of unjust discrimination~\cite[see, e.g.][]{barocas-hardt-narayanan, AClarificationOf, kusner2017counterfactual}. 
This simple example raises the following:

\begin{openpoint}[Sensitive data collection]
\label{a:sensitive_data_collection}
    Should developers of ADM systems keep track of all the sensitive attributes that they would not otherwise record, \textbf{for the sole purpose} of assessing unjust discrimination with respect to those attributes?
\end{openpoint}

According to \citet{buyl2022inherent}, sensitive data collection may not only be morally unacceptable, but also extremely unlikely to occur because people are typically reluctant to share personal information. This could actually pose a significant barrier to the development of ADM systems.

Notice that the issue of sensitive data collection is also explicitly considered by the \citetalias{AIAct}, where it is foreseen that developers of AI systems may process special categories of personal data ``To the extent that it is strictly necessary for the purpose of ensuring bias detection and correction''~\citep[][art~10, 5]{AIAct}. 

\subsection{The problem of aggregation}

Even once we agree on what are the dimensions that contain groups of people to be protected against unjust discrimination, we still have to face the problem of how to aggregate individuals along those dimensions. The simplest example in this respect is that of age. Suppose that we are in a situation in which we agree that there should be no discrimination based on the age of the applicants. We can take the example of credit lending: in the U.S., the \citetalias{ECOA} forbids to discriminate ``on the basis of race, color, religion, national origin, sex or marital status, or age''~\citepalias[][\S1691(a)(1)]{ECOA}.  What do we actually mean by age? Should we consider separate protected groups for each year of age? Or is it enough to aggregate individuals into a handful of broader classes, e.g. $<30$, $[30, 60]$, $>60$? If we opt for the broader classes, how do we select the thresholds? Should we take quantiles of the age distribution of our data, or should we use some common-sense knowledge of how people are actually segmented in the society?

\begin{openpoint}[Group aggregation]
\label{a:group_aggregation}
    The specific identification of most attributes that are commonly considered protected depends on \textbf{alternative ways of aggregating individuals}: what strategy should developers follow to choose the proper aggregation when assessing unjust discrimination? 
\end{openpoint}

This same problem arises for almost all the potentially protected attributes, think e.g. of profession, place of birth, political opinions, disability, ethnicity. Notice that the way in which we cluster individuals into groups may have a significant impact on the assessment of bias and discrimination along that dimension. It may happen, e.g., that individuals below 30 years of age are not discriminated against people over-60, but if you focused on the narrow range of 20-25, individuals in this group may still undergo big disparities compared to people in the range 60-65. 

It is important to note that, at least for some of the potentially protected characteristics, there are concerns and discussions about the prospect of placing people in rigid and exclusive categories~\citep{Lu2022}. For instance, multiracial individuals come from various racial groupings. Indeed, at least on a biological/genetic level, race and ethnicity are now seen as extremely fluid and nuanced ideas rather than simple categorical attributes. Gender and sexual orientation are subject to very comparable criticism.
Despite appearing somewhat abstract, the question of whether it is possible to describe entire groups of individuals based on (roughly stable) human characteristics is a serious one, which we will revisit in \autoref{sec:causality}.

\subsection{How shall we deal with several sensitive attributes \emph{at the same time}?}

Most of the literature on fairness in ML deals with a single sensitive dimension, usually assumed to be binary. Even if this is hardly the case in real-world scenarios, it is true that, when facing with multiple sensitive variables, one could simply repeat the assessment for each of the variables, separately. 

However, this may hide forms of \emph{intersectional bias}~\citep{buolamwini2018gender, roy2023multi}. Namely, the fact that the decision system may have an equal impact on men and women and an equal impact on black individuals and white individuals but still present significant disparities between black women and white men.

This issue arises whenever we assess discrimination with respect to groups: achieving some form of parity at the group level may hide disparities \emph{within} the group, e.g. at the intersection of multiple sensitive characteristics. \citet{kearns2018} call this phenomenon \emph{fairness gerrymandering}.
It is true that this drawback might be easily resolved, in principle, by conducting a fairness evaluation on all potential sensitive group intersections. Unfortunately, this is not a very practical solution, since the number of subgroups to take into account increases exponentially as more dimensions are added, and similarly, the number of data samples in each subgroup rapidly decreases, making it extremely difficult to draw any reliable statistical conclusion~\citep{roy2023multi, kong2022}.

\citet{roy2023multi} further underline that the multi-dimensional aspect of discrimination exacerbates the already mentioned issues of aggregation and categorisation of individuals into groups, and is of course very much connected to the choice of protected groups.

\begin{openpoint}[Intersectional bias]
\label{a:intersectional_bias}
    Is it fair enough to evaluate unjust discrimination with respect to \textbf{sensitive attributes separately}? If not, which \textbf{combinations of sensitive characteristics} should we give priority to (given that we cannot realistically hope to assess all possible combinations)?
\end{openpoint}

\section{What do we mean by unfair discrimination?}
\label{sec:P2}

Clarifying what we mean in concrete terms when we state that a particular group of individuals is discriminated against is probably the single most important piece that is still missing. 
Quantitative studies to assess the presence of unjust discrimination have naturally turned mainly to legal literature and anti-discrimination legislative frameworks to come up with concrete mechanisms underlying unfair practices in decision-making.

The legal debate on the topic, as well as the legislative provisions, is highly diversified, and in various respects still very much open. Moreover, the concept of discrimination has deep roots in both political and moral philosophy, with arguments that go beyond the focus on how to fairly compare people belonging to different social groups, and rather discuss on what grounds individuals should be compared at all, with debates going as far as discussing merit and desert as true drivers of individual success. 

In light of this, it should come as little surprise that the need for fair algorithms has a variety of nuances and complexities, many originating from very fundamental and hotly contested conceptions, which we can barely hope to be met at all.

In the following section, we summarise the basic legal notions that have guided most of the literature on quantitative assessment of unjust discrimination of protected groups in ADM.

\subsection{Direct and indirect discrimination}
\label{sec:direct_and_indirect_discrimination}

The key distinction when analysing the concept of discrimination in legal situations is arguably that between \emph{direct discrimination} and \emph{indirect discrimination}~\citep{sep-discrimination}. 
Roughly speaking, an \emph{indirect discrimination} happens when the decision depends on a sensitive characteristic \emph{through} other variables, therefore not explicitly. While an analysis based on causal reasoning allows to be more specific and detailed ---we refer to Section~\ref{sec:causality} for such a discussion--- an example can provide a straightforward intuition. Suppose you have to develop a decision system to support credit lending decisions based on available applicants' characteristics. Among these characteristics there are, e.g., the level of income, the requested amount, the record of past loans, and the gender attribute. The system will show a \emph{direct discrimination} with respect to gender if it is aware of the gender attribute, and uses it (together with the other variables) to estimate the optimal outcome. For instance, a Machine Learning model that is given access to historical data on loan applications and their (non-)repayment, may learn a statistical association between gender and loan repayments, thus effectively assigning a bonus/malus weight for the \emph{sole gender attribute} of the applicant. 
Conversely, if the system is not aware of the gender attribute, or ---more precisely--- if it does not assign any contribution to the gender attribute itself, the system is not showing a direct discrimination. However, it may nonetheless undergo \emph{indirect discrimination} with respect to gender since it outputs decisions on the basis of the level of income, which is associated to gender. In this last case, the system shows an indirect discrimination with respect to gender, mediated by the level of income.

The direct vs indirect distinction for discrimination is definitely not the only one proposed in the literature or present in legislative frameworks. This approach is mainly found in the European Union and U.K. laws ---see, e.g., \citetalias{Directive200654} and \citetalias{Directive200043}, \citetalias{EqualityAct}, and also~\cite{barnard_hepple_2000}--- while U.S. anti-discrimination laws rely on a similar distinction between \emph{disparate treatment} and \emph{disparate impact}, the former representing forms of explicit discrimination while the latter encompassing a broader set of indirect mechanisms ---see, e.g., \citet{barocas2016big} and \citet[chapter~6]{barocas-hardt-narayanan}.

Perhaps the clearest example of legislation that makes use of direct vs indirect discrimination is once again the \citetalias{EqualityAct}, that gives a  clear definition of direct \citep[Part~2, Chapter~2, Section~13]{EqualityAct} and indirect \citep[Part~2, Chapter~2, Section~19]{EqualityAct} forms of discrimination.

Another well known distinction is the one made within the framework of the theory of \emph{equality of opportunity}, i.e. the idea supporting the need to `level the playing field' for all individuals: different degrees of successes and failures of individuals are fairly distributed only when they ``play'' on a field without ``slopes'' that may advantage some with respect to others.
Experts identify a wide range of possibilities, from a \emph{formal equality of opportunity}, proposing the avoidance of explicit discrimination of protected groups, to forms of more and more \emph{substantive equality of opportunity}, supporting the avoidance of more subtle and indirect mechanisms of discrimination (and sometimes equated to the \emph{equality of outcome}) ---see \citet{barnard_hepple_2000} for a legal perspective and \citet{sep-equal-opportunity} for a philosophical one. Unfortunately, as it is the case for most of the concepts in discrimination theory, the level of consensus about what equality of opportunity actually means in real-world scenarios is incredibly low. Quoting~\citet{sage-equal-opportunity}:
\begin{quoting}
    There is widespread agreement that equality of opportunity is a good thing, even a constituent of a just society, but very little consensus on what it requires. Defenders of equality of opportunity suppose that it requires people to be able to compete on equal terms, on a “level playing field,” but they disagree over what it means to do so. They believe that equality of opportunity is compatible with, and indeed justifies, inequalities of outcome of some sort, but there is considerable disagreement over precisely what degree and kind of inequalities it justifies and how it does so.
\end{quoting}

Against this background, while it is widely recognised that direct discrimination is hardly acceptable, the debate on indirect discrimination is much more nuanced, as we shall also discuss later on (\autoref{sec:business_needs} and, for a causality-based view, \autoref{sec:causality}). Moreover, two factors, among others, contribute to make the distinction at times quite loose and difficult to maintain: 
\begin{enumerate}
    \item a system may not explicitly use sensitive characteristics, but still produce an indirect discrimination through variables that are \emph{per se} not sensitive, but very much associated to sensitive information;
    \item oftentimes, one way to avoid indirect discrimination is to provide favourable conditions to the otherwise unfavoured group. Unfortunately, using dual standards on the basis of sensitive characteristics is precisely what direct discrimination means.
\end{enumerate}
Factor~1 is the well known problem of \emph{proxy variables}, that may be exploited ---possibly with an explicit discriminatory intent--- in order to obtain a discriminatory outcome while maintaining a formal absence of direct discrimination. \citet{bloomberg_amazon} describe the situation of the Amazon same-day delivery service in 6 metropolitan areas in the U.S.: they document a strong correlation between neighbourhoods not reached by the same-day delivery service and the presence of a significant majority of black residents. The most striking example is that of Boston, where the very central neighbourhood of Roxbury (with a 59 percent presence of black residents) was not granted the same-day delivery service, while all the areas surrounding Roxbury are eligible for the service.\footnote{As explicitly declared in \cite{bloomberg_amazon}, Amazon later extended its same-day delivery service to the entire metropolitan area of Boston, as well as to that of New York and Chicago.} Amazon declared that demographics of neighbourhoods played no role in their choices. In cases like this, it is indeed difficult to disentangle direct and indirect forms of discrimination, and agree on what we should and should not tolerate as fair practices. 

The second factor is an example of the so-called \emph{affirmative} (or \emph{positive}) \emph{action}, i.e. providing more favourable conditions to groups of individuals traditionally discriminated against. The other side of the coin is known as \emph{reverse discrimination}, i.e. maintaining dual standards not justified by task-related skills and characteristics~\citep{lipton2018does}.
\citet{kim2022race} refers to~\citetalias{ricci2009}, where a case of affirmative action by the City of New Haven's fire department in favour of black firefighters has been judged to be unfairly discriminating the supposed favoured group (white). The U.S. Supreme Court decision on \citetalias{ricci2009} reports that there was not enough evidence that the performed disparate treatment by the City has avoided a disparate impact:\footnote{Incidentally, in partial support to Section~\ref{sec:causality} on causality, notice that this formulation has a genuine counterfactual nature.} 
\begin{quoting}
    We conclude that race-based action like the City's in this case is impermissible under Title VII unless the employer can demonstrate a strong basis in evidence that, had it not taken the action, it would have been liable under the disparate-impact statute.
\end{quoting}

As noted previously, the landscape is complex and extremely diverse, and a thorough discussion is well beyond the scope of this work. However, we want to make clear that there is no consensus in the scientific communities on what mechanisms actually constitute an unjust discrimination, particularly regarding indirect forms and their potential tension with the direct ones.

In light of this, we state the following:
\begin{openpoint}[Direct and indirect discrimination]
\label{a:direct_and_indirect_discrimination}
    When evaluating unjust discrimination, should developers of ADM systems take into account all the potential direct and indirect ways by which sensitive characteristics may have affected the outcome? Is it acceptable to \textbf{engage in direct discrimination in order to prevent indirect discrimination}?
\end{openpoint}

\subsection{What characteristics represent a source of objective justification for discrimination?}
\label{sec:business_needs}

Open Point~\ref{a:direct_and_indirect_discrimination} is largely intertwined with the discussion on how to deal with indirect discrimination, once we agree that also indirect discrimination may be unacceptable in some cases.  

The focus of the debate is sometimes presented as the identification of `business necessities'. With this term, one loosely refers to the possible presence of characteristics that are fundamentally relevant to the target estimation, therefore their use to perform ADM is justifiable even when such characteristics bear some dependence on sensitive dimensions. Quoting~\citet{barocas2016big}
\begin{quoting}
    This defense [of the presence of a `business necessity'] is, in a very real sense, the crux of disparate impact analysis.
\end{quoting}
The problem is how to decide when a variable is eligible to represent a business necessity.

Incidentally, notice that also direct discrimination may be at times bypassed by business necessities: \citet{romei_ruggieri_2014} suggests the example of the necessity to take explicitly into account gender when advertising for an actor to play, e.g., a male role. This is also explicitly considered in the \citetalias{EqualityAct}, where direct discrimination is always considered unlawful with the exception of direct discrimination with respect to age, that is instead permitted provided that objective justification is given \citep[Part 2, Chapter 2, Section 13]{EqualityAct}.  

Legislative frameworks make explicit reference to this possibility. For instance the \citetalias{Directive200654} on non-discrimination of men and women in matters of employment and occupation, has provisions of objective justification already in the very definition of indirect discrimination (art~2,1(b)) 
\begin{quoting}
    ‘indirect discrimination’: where an apparently neutral provision, criterion or practice would put persons of one sex at a particular disadvantage compared with persons of the other sex, unless that provision, criterion or practice is objectively justified by a legitimate aim, and the means of achieving that aim are appropriate and necessary
\end{quoting}
The key ingredients to have an objective justification for indirect discrimination are here identified to be a ``legitimate aim'', and ``appropriate and necessary means'' to achieve it.

The \citetalias{EqualityAct} defines indirect discrimination as follows \citep[Part~2, Chapter~2, Section~19]{EqualityAct}:
\begin{quoting}
    \begin{itemize}
        \item[(1)] A person (A) discriminates against another (B) if A applies to B a provision, criterion or practice which is discriminatory in relation to a relevant protected characteristic of B's.
        \item[(2)] For the purposes of subsection (1), a provision, criterion or practice is discriminatory in relation to a relevant protected characteristic of B's if—
        \begin{itemize}
            \item[(a)] A applies, or would apply, it to persons with whom B does not share the characteristic,
            \item[(b)] it puts, or would put, persons with whom B shares the characteristic at a particular disadvantage when compared with persons with whom B does not share it,
            \item[(c)] it puts, or would put, B at that disadvantage, and
            \item[(d)] A cannot show it to be a proportionate means of achieving a legitimate aim.            
        \end{itemize}
    \end{itemize}
\end{quoting}
Again, making explicit reference to a ``legitimate aim'' that can justify the use of variables that are associated with protected characteristics.

Similar arguments can be found in U.S. laws. The \citetalias[Title~VII]{USCivilRights} on non-discrimination in the workplace, among the criteria in support of the presence of unjust discrimination, has the following: 
\begin{quoting}
    a complaining party demonstrates that a respondent uses a particular employment practice that causes a disparate impact on the basis of race, color, religion, sex, or national origin and the respondent fails to demonstrate that the challenged practice is job related for the position in question and consistent with business necessity
\end{quoting}
Here ``job-relatedness'' and ``business necessity'' are cited as legitimate reasons to have indirect discrimination.

\cite{barocas2016big} discuss in details the evolution and oscillation of the interpretation of such terminology by the U.S. Supreme Court, somehow emphasising the ambiguity of it. The point is that all these concepts are inherently fuzzy. One may interpret a characteristic to be job related if it correlates with job performance, and similarly job performance may be considered a legitimate aim. 

\cite{albach2021role} present the results of extensive surveys (2157 Amazon Mechanical Turk workers) on different domains to understand what are the \emph{perceived} drivers of fairness in Machine Learning applications. Interestingly, they find that `relevance' and `increases accuracy' are the 2 dimensions that explain most of the perceived fairness of a given variable. More precisely, in a first study, the attribute `increases accuracy' was not among the exposed ones, and `relevance' was by far the most important dimension explaining perceived fairness. Acknowledging the possible ambiguity of the term `relevance', the authors designed further studies, either replacing `relevance' with `increases accuracy', or adding `increases accuracy' to the possible drivers. They conclude that most of the times the survey participants perceive `relevance' as a placeholder for the characteristic of increasing performance, and that this single dimension contributes the most to the perceived fairness of a given variable. 

These results seem to support the loose interpretation of business necessities as any variable that can increase predictive performance.    
But notice that this interpretation strongly undermines the very concept of indirect discrimination: if all the variables that correlate with the outcome can be used legitimately irrespective of their dependence on sensitive characteristics, then we hardly have indirect discrimination at all.

Against this background, practitioners who need to assess and possibly mitigate any unjust discrimination when developing an ADM system face the following:

\begin{openpoint}[Objective justification]
\label{a:business-needs}
    What qualities should an attribute have, if any, to be \textbf{eligible for use} in automatic judgements, even if it serves as a basis for indirect discrimination? \footnote{It should be noted that we are slightly abusing terminology here, as indirect discrimination is typically regarded as occurring when such an objective justification cannot be provided.}
\end{openpoint}

\paragraph{Long-term impact} A very subtle issue is what happens on the long-term given some notion of unjust discrimination. It is not at all clear that fair ADM systems are going to push society towards an actual equality of opportunity. Suppose we acknowledge that income is allocated on average unevenly between men and women, and that this is due to historical bias that, as a society, we aim to eradicate. Should fair ADM systems avoid gender disparities due to income in their outcomes? This would mean that ADM systems should favour women over men, at the same level of income. In the long run, is this aligned with the overall goal of eradicating gender bias? Needless to say, the answer to this question is: we don't have a clue.  Indeed, a broad view of equality of opportunity~\citep{barocas-hardt-narayanan} is not focused on ADM at all, but is much more top-down and global, discussing how society overall should be designed, how its institutions should function, what are the incentives that, in the long run, would tend to actually level the playing field for every individual, without the need of any ``local'' intervention on single ADM systems.  
As an example of this broader spirit, we like to cite the article~3 of the~\citetalias{ITConst}, which reads:
\begin{quoting}
    It is the duty of the Republic to remove those obstacles of an economic or social nature which constrain the freedom and equality of citizens, thereby impeding the full development of the human person and the effective participation of all workers in the political, economic and social organisation of the country.
\end{quoting}

\subsection{The risk of over-reliance on fairness metrics}
\label{sec:overreliance}

Sometimes, the various forms of unjust discrimination that we have encountered so far are associated with specific \emph{observational metrics}. With this term, we mean quantities that can be computed by having access to samples of the joint distribution $P(X, A, Y, S)$, where variable $A$ represents the sensitive attribute, $X$ denotes the set of other (non-sensitive) characteristics, $Y$ is the target variable, and $S$ is the model outcome.

A discussion of fairness metrics is out of scope in this study ---there is plenty of literature on the topic \cite[see, e.g.,][and references therein]{AClarificationOf, mitchell2021algo}--- we here just recall the main classes of observational metrics. In general, observational metrics are expressed as statistical relationships among (some of) the variables introduced above (indeed, they are sometimes called \emph{statistical metrics}). In particular, we can distinguish the following classes of observational metrics~\citep{barocas-hardt-narayanan}:
\begin{itemize}
    \item \textbf{independence}: $S \indep A$;
    \item \textbf{conditional independence}: $S \indep A \mid X'$, with $X' \subseteq X$;
    \item \textbf{separation}: $S \indep A \mid Y$;
    \item \textbf{sufficiency}: $Y \indep A \mid S$.
\end{itemize}

Independence is usually associated with disparate impact: the fact that the model outcome bears no dependence on the sensitive attribute, directly translates into a substantial parity in the outcomes. In the simple case of binary classifications, where $S \in\{0, 1\}$,\footnote{More precisely, most binary classifiers outputs a real-valued \emph{score}, which is an estimate of the probability of the input belonging to the positive class. In those cases, the binary outcome is a consequence of thresholding the score.} independence is indeed equivalent to what is known as \emph{Demographic Parity} (or \emph{Statistical Parity}), namely the equality of acceptance rate among all demographic groups along the sensitive dimension.

The introduction of business needs is sometimes associated with conditional independence and sometimes with separation. The idea behind these associations is the following: if we agree that some variables (that we collectively label with $X' \subseteq X$) are acceptable grounds to make decisions irrespective of their dependence on sensitive attributes (i.e., they represent business necessities), then the natural quantity to inspect is the independence of the outcome on sensitive attributes \emph{stratified} by those variables. Namely, if we agree that the use of the level of income for credit decisions is justified by business needs, then we may require to have equality of acceptance rate between, say, male and female applicants \emph{within the same level of income}. In other words, we are requiring that the only gender disparity we are willing to tolerate is the one justified by the level of income. In a seminal paper on Conditional Demographic Parity, \citet{WACHTER2021why} suggest the use of this type of metric in relation to the EU non-discrimination legal framework. 

Similarly, in the case of separation metrics, we allow only disparities justified by the target variable: e.g., we tolerate gender disparities as long as they are justified by actual disparities in repayment rates.
This is in line with the previous discussion on business necessities (\autoref{sec:business_needs}): if business needs can be justified on grounds of evidence of predictive performance, then metrics of the separation class are indeed more appropriate.

Notice, incidentally, that the set of variables $X'$ used as conditions determines how much the corresponding independence statement is focusing on smaller and smaller groups. Fair-AI literature discusses the distinction between \emph{individual} and \emph{group} metrics, depending on the granularity of the observations compared: while group metrics are computed as averages over groups that are then compared to one another, individual metrics consider comparison among individuals. 
At the level of observational metrics, this distinction is somehow overlapping to that between direct and indirect discrimination: namely, the more granular we push the metrics the more we concentrate on direct effects, and vice versa~\citep{AClarificationOf}. We shall see in \autoref{sec:causality} that these relationships can indeed be decoupled with the lens of causal reasoning. To the best of our knowledge, this subtlety is absent in most legal debates.

While these arguments are valid, they unfortunately conceal certain possible weaknesses that we will cover in the reminder of the section.

\paragraph{The label problem} Many of the statistical metrics (namely, those of the separation and sufficiency classes) rely on the comparison between model outcomes ($S$) and target labels ($Y$). Unfortunately, relying on labels is delicate for (at least) two types of drawbacks. On the one hand, we have a problem of actual \emph{access} to labels: e.g., when granting loans, there is no information about applicants that do not receive loans in the first place (namely, access to $Y$ is possible only for applicants with positive outcome $S$). Moreover, even when we do have access to labels, oftentimes there may be a significant delay, as it is the case for repayment of loans, or for job performance in recruiting decision-making: in such scenarios we can monitor labels (and thus fairness metrics based on those labels) only considerably \emph{ex-post}. \footnote{Of course, this \emph{ex-post} effect is there also when monitoring algorithmic performance with respect to such targets.} 
On the other hand, there may be an issue of \emph{label bias}~\citep{obermeyer2019dissecting}: this happens when seemingly effective proxies for ground truth are chosen as target, but there are mechanisms (often involved and ignored by the developer and/or the decision-maker) by which those proxies embed forms of bias. Incidentally, notice that the issue of a sound choice of the target variables in fair-AI goes even beyond the choice of a statistical metric potentially grounded on biased labels: optimising an ADM system by learning from biased labels is of course problematic under multiple perspectives~\citep{biasondemand}.

\paragraph{The incompatibility problem} It is well-known that statistical metrics are generally not mutually compatible~\citep{kleinberg2016inherent, barocas-hardt-narayanan}. For example, one cannot hope to have an ADM system presenting no Demographic Disparities and, at the same time, having equal error rates for different (sensitive) groups of individuals. This is true for all the classes of metrics listed above, a part for degenerate cases.
This evidence implies that, when using such metrics, one must make choices, that are ultimately associated with the perspective one has on unjust discrimination. 
Some works have proposed guidelines ---usually in the form of decision trees or diagrams--- to help finding the most appropriate statistical metric given domain-specific constraints~\citep[see,~e.g.,][]{makhlouf2021ontheapplicability, makhlouf2021machinelearning, haeri2022promises, smith2023scopingfairness}. However, as the authors of such works clearly acknowledge, the process of following the proposed decision diagrams is itself complicated, involving necessarily multi-disciplinary competencies, and in any case they warn not to take these diagrams too categorically or as a set of well-established prescriptions. 
To make things even more blurry, if it is true that having perfect parity with respect to different metric classes is mathematically impossible, allowing a limited level of disparity may be attainable with multiple metrics at the same time~\citep{bell2023thepossibility}, suggesting that focusing too much on a single metric maybe counter-productive after all. 

\paragraph{The threshold problem} Once practitioners have computed the values of observational metrics, they face the problem of ``deciding'' whether the numbers found are enough to raise a warning or not. A 10\% difference in acceptance rate between male and female is fair enough? What about using the ratio of acceptance rates instead of the difference? Borrowing from~\citet{Ruggieri_Alvarez_Pugnana_State_Turini_2023} ``The apparently innocuous choice between the algebraic operators (1) or (2) [difference vs ratio], however, has an enormous impact on how decisions are affected.''   
The threshold problem is as trivial as it sounds: statistical metrics are real-valued measures of (some notion of) disparities, they are not a binary trigger. 
Some attempts to face this issue can be found in the U.S. context. One is the well-known \emph{four-fifths rule}~\citep{fourfifth}, stating that an acceptance rate (in recruiting setting) of the `unfavoured' group less than 80\% of the acceptance rate of the `favoured' group constitutes an evidence of disparate impact.\footnote{Incidentally, notice that this is a rule on the \emph{ratio} of acceptance rates.} To be clear, the \citet{fourfifth} adds several \emph{caveats} to that rule, that are often overlooked, e.g. that a greater ratio may still constitute an adverse impact ``when significant in both statistical and practical terms'', while smaller ratios may not be enough evidence for adverse impact, e.g. when they are based on small samples. Other approaches rely on statistical significance of proportions: we can mention the \emph{Castaneda rule}~\citep{romei_ruggieri_2014}, named after the \citetalias{castaneda1977} law case, which prescribes a significance test on $P(A=1 \mid S=1)$. More precisely, if the proportion of protected individuals selected ($P(A=1 \mid S=1)$) departs more than ``two or three'' (\emph{sic.}) standard deviations from what we expect in the independence scenario ($P(A=1) \times P(S=1)$), then a significant disparity is considered at play.\footnote{In \citetalias{castaneda1977} a binomial distribution is assumed for the proportion $P(A=1 \mid S=1)$. Generally speaking, considering a normal approximation to the binomial proportion, 2 (3) standard deviations correspond to a 95\% (99.7\%) confidence interval. In other words, a disparity is assumed whenever the observed proportion would be observed less than 5\% (0.03\%) of the times if no real disparity was present.} 

In general, developers of ADM systems are therefore challenged by the following:
\begin{openpoint}[The threshold problem]
\label{a:threshold}
    Given an observational metric, on what grounds should we decide \textbf{how small the measured disparity should be} in order to be considered fair enough?
\end{openpoint}

\paragraph{Lack of causality} Another drawback of purely observational metrics is that they are blind with respect to the underlying generation mechanisms: it is well known that statistical correlations among variables may have very different natures. They may be due to a direct causation of one variable on another; to causation mediated by other variables; or may be spurious, i.e. due to some common factor causing both of them; or may even be spuriously created when conditioning on a common effect~\citep{Peters, neal2020introduction}. More details on the topic of causality in fair-AI are discussed in \autoref{sec:causality}.

A vivid critic of the ``metric approach'' to fairness can be found in \cite{selbst2019}, where the authors face the fair-AI problem from the perspective of Science and Technology Studies. In particular, they talk about a \emph{Formalism Trap}: fairness and discrimination are complex concepts, that contain many subtle dimensions, among which a \emph{procedural}, a \emph{contextual}, and a \emph{contestable} dimension, that cannot be grasped by relatively simple statistical quantities. Of course, this does not imply that statistical measures are altogether wrong or useless, but only that they must be utilised as tools, to be set in a wider context.  

Indeed, survey studies about fairness \emph{perception} in ML show a very diverse and nuanced landscape, where perspectives on fairness are associated with socio-demographic factors of the participants~\citep{grgichlaca2022dimensions}, and consensus on fairness notions is very low~\citep{starke2022fairnessperceptions, harrison2020anempirical}. 

Overall, considering together all these aspects about observational metrics, we can state the following:
\begin{openpoint}[Over-reliance on observational metrics]
\label{ap:overreliance_on_observational_metrics}
    Purely observational fairness metrics should be taken with a grain of salt. At best, they can  be used \textbf{as a means for a deeper reasoning} on the mechanisms underlying a phenomenon, rather than a final word on the presence or lack of unjust discrimination. A clear connection between quantitative metrics and unjust discrimination is still missing. 
\end{openpoint}

\section{The importance of causality}
\label{sec:causality}

Legal provisions against discrimination frequently have a counterfactual bent, either overtly or covertly.

Regarding direct discrimination, the \citetalias{Directive200654} has the following definition: 
\begin{quoting}
    where one person is treated less favourably on grounds of sex than another is, has been or would be treated in a comparable situation
\end{quoting}
while, in \citetalias{carson1996}, we read:
\begin{quoting}
    The central question in any employment-discrimination case is whether the employer would have taken the same action had the employee been of a different race (age, sex, religion, national origin, etc.) and everything else had remained the same. 
\end{quoting}

Regarding indirect discrimination, the 2015 U.S. Supreme Court decision for \citetalias{texdept2015} established a ``robust causality'' requirement for the plaintiff to demonstrate the presence of disparate impact.  

In general, it is reasonable to claim that, to inspect whether or not an unjust discrimination is present, we need to understand what are the causal relationship among relevant variables and sensitive ones. E.g., how do we disentangle direct and indirect discrimination? How do we know that a variable that we can argue to be necessary for our business is the one responsible for the observed disparate impact?

\begin{figure}
\begin{center}
\begin{tikzpicture}
    \node[state] (d) at (0, 1.5) {$D$};
    \node[state] (g) at (-2, 0) {$G$};
    \node[state] (y) at (2, 0) {$Y$};
    \draw[dashed] (g) -> (y)  node [midway] {?};
    \path (g) edge (d);
    \path (d) edge (y);
\end{tikzpicture}
\end{center}
\caption{\textbf{1973 Berkeley graduate admission.} $G$ represents gender, $D$ the chosen Department and $Y$ the admission outcome. The dashed line represents the possible direct effect of gender on the admission choice.}
\label{fig:berkeley}
\end{figure}
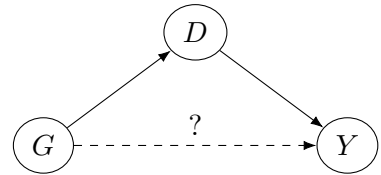

A well-known and extensively studied example of the importance of causal reasoning to assess fairness in decision-making is that of the graduate admissions at Berkeley in 1973 as discussed by~\citet{bickel1975sex}. Historical data aggregated over the six largest departments showed a significant face-value disparity of admission rate between male and female applicants, with male being accepted with a rate 14 percentage points higher (44\% vs 30\%).  This seemingly huge discrepancy in favour of men applicants turned out to be reversed in 4 out of 6 departments, when inspecting admission rates data for each of the 6 department separately. This is an example of the so-called \emph{Simpson's paradox}, the well known fact in probability and statistics by which aggregate trends may disappear or revert at a more granular level, due to relative frequencies of individuals in subpopulations. In the Berkeley admission case, men applicants were much more numerous in departments where the admission rate was higher (both for men and women), while female applicants concentrated more on departments with lower acceptance rate.

The bottom line\footnote{First drawn by \citet{bickel1975sex} and then deeply discussed by Judea Pearl; see e.g. the popular audience book by~\citet{pearl2018book}.} is that, when conditioned on department choice, the observed gender disparity disappears, revealing the absence of a direct discrimination by admission committees. The observed disparity can be seen as an indirect effect due to the fact that female applicants tended to choose more competitive departments. In the language of causal theory, department choice is a \emph{mediator} between gender and admission (Figure~\ref{fig:berkeley}). 

We refer to \cite{barocas-hardt-narayanan} and \cite{plecko2022fairness} for an in-depth analysis of this exemplar case,
we here simply rephrase what is beautifully reported in \citet{barocas-hardt-narayanan}, namely that even when the causal structure is revealed and we know the mechanism by which the disparity among genders appears on the observational data, the question about discrimination is still there, we have merely scratched the surface. Namely, one could (and should?) ask and investigate the reasons why women tended to apply to more competitive departments in the first place. 
In fact, what the causal structure has revealed is that there is no direct discrimination, and that the entire observed disparity is a consequence of the choice of the department by the applicants. But the causal structure cannot say whether this channel of disparity is a fair one or not. Namely, we are back at the discussions of sections~\ref{sec:direct_and_indirect_discrimination}-\ref{sec:business_needs} and to Open Points~\ref{a:direct_and_indirect_discrimination}-\ref{a:business-needs}.

It is interesting to underline that, even when employing observational metrics, the process of choosing which of the many available metrics are the most appropriate for a particular setting is a form of causal reasoning in disguise. Indeed, in Open Point~\ref{ap:overreliance_on_observational_metrics} we suggest to use observational metrics precisely as a means to understand the mechanisms underlying a specific observational distribution. As hinted in \autoref{sec:overreliance}, metrics based on conditional disparities are appropriate when we assume the presence of legitimate mediators: conditioning on a mediator is precisely the way through which we can isolate other unwanted sources of disparities, analogously to the 1973 Berkeley admission case.

In some sense, the causal framework allows us to capture more clearly what we should compare in order to assess a specific discrimination mechanism. Analogously to the well-known Fairness Through Awareness approach \citep{dwork2012fair}, that suggests to compare observed individuals on the basis of a to-be-defined similarity metric, the causal framework suggests to compare an observed individual to a counterfactual self for whom sensitive conditions are changed in a ``controlled'' way. Depending on the ``causal channels'' that we block or let free, we can pinpoint different discrimination mechanisms. 

In the Berkeley graduate admission example we may flip gender \emph{while keeping the department fixed}, thus measuring the direct effect only. Or we could flip gender and \emph{let the department free to vary}:  
a male transported in the counterfactual world where he is a woman would have himself chosen higher demanding departments, thus this feature would change as well. In this case we are targeting both direct and indirect effects, the latter channeled through the department choice.

\begin{figure}
\begin{center}
\begin{tikzpicture}
    \node[state] (z) at (0, 1.5) {$Z$};
    \node[state] (a) at (-2, 0) {$A$};
    \node[state] (w) at (0, -1.5) {$W$};
    \node[state] (y) at (2, 0) {$Y$};
    \path (a) edge (y);
    \path (a) edge (w);
    \path (w) edge (y);
    \path (z) edge (w);
    \path (z) edge (y);
    \path[bidirected, bend right=40] (z) edge (a);
\end{tikzpicture}
\end{center}
\caption{\textbf{Standard Fairness Model}~\citep{plecko2022fairness}. The bidirected dashed edge represents the possible presence of hidden confounding.}
\label{fig:SFM}
\end{figure}
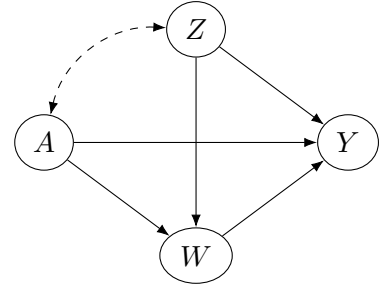

Building on seminal works on causality in fair-AI, such as those by \cite{kusner2017counterfactual} and \cite{Chiappa_2019}, \cite{plecko2022fairness} perform a remarkable analysis of fair-AI through the lens of causal theory. They suggest a basic causal graph (Figure~\ref{fig:SFM}), that they call Standard Fairness Model, containing 4 categories of nodes: the sensitive attribute $A$, the target variable $Y$, mediator variables $W$ and other non-mediator variables $Z$. Variables $Z$ can have impact on $Y$, and they can be confounded with $A$, meaning that there may be unobserved common causes to $Z$ and $A$. The Boston same-day delivery example discussed in \autoref{sec:direct_and_indirect_discrimination} can be seen as an example of confounding, where ethnicity of residents ($A$) has some ancestor cause in common with the neighbourhood they end up living in ($Z$).  
Given this basic structure, they identify 3 building blocks of the unconditional dependence of $Y$ on $A$:
\begin{enumerate}[a)]
    \item \emph{Direct Effects}: represented by the direct arrow from $A$ to $Y$, measurable by the variation in $Y$ resulting from flipping the attribute $A$ while keeping $Z$ and $W$ fixed;
    \item \emph{Indirect Effects}: represented by arrows going through $W$, measurable by the variation in $Y$ resulting from the flipping $A$ and letting $W$ vary, but keeping $Z$ and $A$ fixed;
    \item \emph{Spurious Effects}: represented by the confounding mechanism through $Z$, measurable by the variation in $Y$ for different \emph{observed} values of $A$, without considering the causal effects of $A$.
\end{enumerate}

For the mathematically inclined, these three types of effect read:
\begin{subequations}
\begin{align}
    \text{DE}_{a_0, a_1; X'} = & P(Y_{a_1, W_{a_0}} \mid X') - P(Y_{a_0} \mid X'), \label{eq:de}\\
    \text{IE}_{a_0, a_1; X'} = & P(Y_{a_0, W_{a_1}} \mid X') - P(Y_{a_0} \mid X'), \label{eq:ie}\\
    \text{SE}_{a_0, a_1; X'} = & P(Y_{a_0} \mid a_1, X') - P(Y_{a_0} \mid a_0, X');\label{eq:se}
\end{align}
\label{eq:causal_effects}
\end{subequations}
where, e.g., the expression $Y_{a_1, W_{a_0}}$ denotes the counterfactual distribution of $Y$ when $A$ is set to $a_1$ but $W$ is kept at the value it would have taken when $A$ is set to $a_0$.
We have here denoted with $X'$ a generic set of variables, it could be, e.g., $A$ itself, or $Z$, etc. The conditioning on $X'$ deserves some additional explanation.  Recall that the conditioning event represents the subset of individuals that happen to meet the given characteristics: conditioning on $A=a_0$ means to restrict the attention to individuals that we observe having value $a_0$ for $A$. As mentioned in \autoref{sec:overreliance}, the more we narrow down with the conditioning event, the more the effect we are capturing is individual in nature. If we stick to pure observational distributions, this concept remains somewhat ambiguous, since the type of the effect and the conditioning are entangled to one another. With the tools of causal analysis things become much more transparent: on one side we have the type of mechanism we are targeting, on an independent level we have the resolution with which we are looking at those mechanisms. The two extremes are: 
\begin{enumerate}[i)]
    \item the individual level, or \emph{unit} level, where we have a measure of the effect for each single individual;
    \item the unconditional level, or \emph{population} level, where we have a single measure representing the overall average effect.
\end{enumerate}
\cite{plecko2022fairness} call the individual level metrics \emph{more powerful}, since the larger the granularity, the more we can average out effects that are nevertheless present at a finer-grained level. Indeed, whenever we measure a null effect by equations~\eqref{eq:causal_effects}, we can only say that there is no discrimination at the level $X'$, but we cannot draw any conclusions on finer levels. On the contrary, if we get a non-zero metric we can be sure that some discrimination is at play.
Unfortunately, the finer-grained we aim, the higher the chance that we cannot compute the effect on the basis of observational data alone, and we need to add further modelling assumptions ---an issue known as \emph{identifiability} of causal effects~\citep[for further details, see][]{plecko2022fairness}.
In particular, individual effects are intrinsically not computable via observable data alone, since an individual cannot be \emph{observed} into distinct states. In contrast, if we average at a coarser-grained level, we can compare similar individuals (i.e., sharing the same values of the covariates $X'$ that identify the desired level) \emph{observed} into different treatment states.

Notice that spurious effects lie at a different level with respect to direct and indirect ones: while the latter are variations in outcome \emph{in response} to intervention on $A$ and/or its descendants, spurious effects are not causally \emph{after} $A$, but \emph{before} it. This is why Eq.~\eqref{eq:se} is the only one in which the change in $A$ is \emph{observed} and not due to intervention. Moreoveor, to isolate the effect, the causal consequences of a change in $A$ are suppressed by measuring the counterfactual distribution of $Y_{a_0}$.
Drawing again from the Boston same-day delivery example, if we intervened on ethnicity we would have no consequences on having/not-having access to the same-day delivery service. To measure the dependence we have to compare an observed black individual (say $A=a_0$) to an observed white individual ($A=a_1$) in the counterfactual setting where she is black ($Y_{a_0}$).   

We have here focused on the simple causal structure presented in Figure~\ref{fig:SFM}, but of course things can be much more involved than that. In fact, \cite{plecko2022fairness} extend their analysis by ``unpacking'' the confounder and mediator variables ($Z$ and $W$, respectively), therefore providing an exhaustive framework including cases when $Z$ and $W$ are multiple, and also accounting for the possibility that any of those variables represent a business necessity.

With this concise overview of causality-based fair-AI, we hope to have made two points clear: \emph{i)} that, on the one hand, the causal framework is a tool in and of itself for analysing the phenomenon under study in greater detail, and \emph{ii)} that, on the other hand, going deeper often reveals richer and more complex structures, as we have seen with the introduction of spurious effects or the decoupling of individual-vs-population level and the mechanisms type level.

Overall, we encourage a discussion on the variables at play and their particular role before even thinking about any fairness metric to be computed and monitored.

\subsection{Limits of causality-based approaches}
\label{sec:causality_limits}

Unfortunately, causality-aware approaches come with plenty of drawbacks.

One of the problems, that we briefly mentioned above, is that some of the more individual-level discrimination effects are not computable with the sole observational distribution and the (assumed) causal structure (\emph{identifiability} problem): the functional relationships among the variables are needed, the so-called \emph{structural equations}. Relying on structural equations introduces an additional level of assumptions and estimations beyond those of the graph structure~\citep[see, e.g.,][]{Peters, neal2020introduction}. 

In general, causal theory requires rather strong assumptions, some of which are not verifiable. Blind computation of causal effects when assumptions do not hold can lead to false results, and therefore to risk of harmful actions in response to such results. 
\cite{fawkes2022selection} argue that the usual assumptions that underlie causal analysis are not really realistic when dealing with fairness, and call for more care when assuming the underlying class of causal models.

\begin{openpoint}[Assumptions of causal structure]
\label{ap:assumptions_of_causal_structure}
    Causal tools require strong, often unverifiable, assumptions. \textbf{Downstream consequences of wrong assumptions can lead to wrong or even harmful actions}. Therefore, particular care must be taken when relying on such tools. 
\end{openpoint}

While Open Point~\ref{ap:assumptions_of_causal_structure} calls for care in the assumptions underlying causal relationships, 
at an even more fundamental level, some scholars have raised concerns regarding causal graph \emph{nodes}. In particular, about the very possibility of using human attributes as nodes in a causal graph. 
The point is that, when we make causal statements about human attributes, oftentimes the direction of causality is far from being obvious. This ambiguity is beyond the ``standard'' difficulty of causality assumptions that we have mentioned above: it is rooted in the very meaning of human attributes.
\cite{hu2020s} make the example of \emph{religious belief}: in particular, they mention the fact that `being catholic' can be seen both as a cause and as a consequence of a lot of different characteristics, such as `believe in the resurrection of Christ', or `believe in saints'.  
\cite{barocas-hardt-narayanan} also make a similar example: we can reasonably assume that one's religious affiliation may causally influence her/his education, thus seeing education as a consequence of religious belief. But we can just as well agree on the opposite: the level of education has impacts on one's spiritual attitudes and believes. The point is that education and religious believes are too intertwined to be cast in a simple cause-effect relationship. 
\cite{hu2020s} talk about \emph{constitutive} relationship as opposed to \emph{causal} relationship. A constitutive relationship indicates the proprieties in virtue of which an object ---in our case a human attribute--- is what it is. In this view, the propriety of `believing in the resurrection of Christ' is neither a cause nor causes that of `being catholic'. It rather constitutes it along with many other proprieties bundled together (e.g., the duty of going to mass, the infallibility of the Pope). By that it is meant that without this specific belief (and other rooted practices), the meaning of the social category `catholic' would be totally different. If this ontology of human attributes is correct, what insight can we gain for causal approaches to fairness? A first suggestion is that the difficulties in drawing causal arrows come from the attempt to provide causal interpretation of a non-causal relation. Following \cite{hu2020s}, we understand that constitutive proprieties play an important role in cases of discrimination based on sensitive attributes. Bearing some physical traits can rarely by itself justify discriminatory outcome. In most cases unfair outcome will be motivated by inter-subjective believes and practices that are constitutive proprieties of sensitive attribute as social kind. While rather informative, these remarks poses a serious question to causal approaches in fairness and may encourage toward more encompassing methods of modelling inferences.

Among the experts in causal theory, one school of thought adheres to the well-known motto ``no causation without manipulation''~\citep{holland1986}: to draw causal inference on something, we should be able to manipulate it. Following this perspective, a notable proposal is that of shifting from modelling an attribute to modelling the \emph{exposure to attribute}~\citep{Rubin2011}. In practical terms, this implies exposing someone to something that is manipulable and that represents a given attribute, such as the \emph{declared} name, age, and other features on a CV. We can easily think of manipulating (i.e., faking) information declared on a CV, without the need of manipulating the actual characteristics of the individual.
Unfortunately, even if useful and effective in some cases~\citep{bertrand2004are}, this also comes with drawbacks. In order for this strategy to work, it appears that we need to model the causal implications of a change in a declared attribute, thus understanding all the different perceptions that an attribute could possibly generate~\citep{barocas-hardt-narayanan}. Ultimately, there is no way of coming up with such a model without an understanding of the social perception and meaning of the actual attribute and category, that is the thing that raised the doubts to begin with. In other words, modelling perception to sensitive attributes is as problematic as modelling the attributes themselves. This aligns with the alternative view of causality as the identification of a \emph{generative process}, rather than relying only on direct manipulation~\citep{goldhorpe2001causation}. Moreover, it appears that intervening on exposure to attributes triggers different downward changes. Considering the example of declared origin on a CV, if we tweak the declared origin we are acting on the social perception of that attribute, but we miss all the (causal) connections that would be present if we tweaked the true origin (whatever this may mean), e.g. education, and so forth. These difficulties may unveil a forced commitment to some constructivist theories of human attributes~\citep{mallon2016construction}.

In general, the problem of causality in social studies is very much debated, several alternative definitions and mechanisms have been proposed and analysed, but there is no clear consensus in the scientific community~\citep{hirschman2014formationstories, goldhorpe2001causation}, which calls for care in handling such concepts. 
The inherent conclusion is that causal modelling and related discrimination issues vitally depend on our ability to understand the relationships between human attributes, users' perception of them, and the representational power of the objects we choose as nodes in the graph.  

We can summarise these arguments by stating the following:
\begin{openpoint}[Use of human attributes as causes]
\label{ap:use_of_human_attributes_as_causes}
    If it is true that causal mechanisms underpinning phenomena are paramount to evaluate the presence of possible unjust discrimination, a particular \textbf{care must be taken on the meaning we assign to nodes} representing human attributes and social categories. 
\end{openpoint}

\paragraph{Remark} By raising Open Points~\ref{ap:assumptions_of_causal_structure}-\ref{ap:use_of_human_attributes_as_causes} we are by no means supporting the idea that causal theory shouldn't be used any longer because there are too many inherent dangers and drawbacks. Quite the opposite, we promote the use of causal tools, and recognise the value in such approaches, but we also call for an awareness of their weak points and limitations.

\section{Conclusions}

In this work, we support the view that the understanding of the landscape of unjust discrimination in ADM, despite the impressive work done in the last decade or so by the fair-AI community, is not yet mature enough to be ``put into practice''. 

We believe that the requirement to develop fair algorithms is still too vague and that, in order to be put in place, we have to clarify and be aware of a set of open points, most of which are societal in nature rather than technical. Given the strong multidisciplinary content, this challenge will be best addressed and discussed when researchers and practitioners from the various fields involved (e.g., statisticians, AI experts, ethicists, and legal experts) collaborate toward the goal, potentially creating a shared vocabulary and set of working notions. In fact, this is the spirit that also guided this work.

As a positive example of legislative framework, we would like to mention the \citetalias{EqualityAct}, that takes an organic perspective on anti-discrimination, clearly stating what are the characteristics that should be considered protected in any domain, and also taking the position of prohibiting any form of direct discrimination. 

As a final remark, notice that there are of course other problematic aspects of unjust discrimination in ADM systems that are somehow out of the scope of this work and would have required a much broader analysis. We can cite, e.g., how to face the challenge of bias in models with unstructured data such as images and text, especially in generative AI models such as the modern Large Language Models; or more technological problems, such as those raised by the modularity of AI systems, that are usually composed of several steps and components, making it complicated to clarify how fairness issues may propagate through the process.

\section*{Author contributions}
D.R. conceived the main ideas presented and outlined a first draft of the manuscript. D.R. and A.C. contributed to the overall structure of the manuscript. G.N. contributed, in particular, in shaping and writing the section on limitations of causal theory. All authors contributed to the discussion of the ideas presented and to the overall review of the manuscript.

\bibliography{references.bib}


\end{document}